\documentclass[]{llncs}
\usepackage[T1]{fontenc}
\usepackage{makeidx}  % allows for indexgeneration
\usepackage{etoolbox} % for "\patchcmd" macro

\usepackage{subcaption}

\usepackage{multirow}

\usepackage[utf8]{inputenc}

\makeatletter
\patchcmd{\ps@headings}{\rlap{\thepage}}{}{}{}
\patchcmd{\ps@headings}{\llap{\thepage}}{}{}{}
\makeatother
\usepackage{graphics}           
\usepackage{times}              
\usepackage{amsmath}            
\usepackage{amssymb}            
\usepackage{graphicx}
\usepackage{algorithm}
\usepackage[noend]{algpseudocode}
\usepackage{booktabs}
\usepackage{color}
\usepackage[nocompress]{cite} % sorts citations automatically, e.g., "[1],[2],[3]" instead of "[2],[3],[1]".
\definecolor{instructioncolor}{rgb}{.5,.5,.5}

\usepackage[font=small]{caption}

\def\eqref#1{Eq.~(\ref{#1})}

\makeatletter
\usepackage{xspace}
\DeclareRobustCommand\onedot{\futurelet\@let@token\@onedot}
\def\@onedot{\ifx\@let@token.\else.\null\fi\xspace}

\makeatother

\usepackage{array}
\newcolumntype{L}[1]{>{\raggedright\let\newline\\\arraybackslash\hspace{0pt}}m{#1}}
\newcolumntype{C}[1]{>{\centering\let\newline\\\arraybackslash\hspace{0pt}}m{#1}}
\newcolumntype{R}[1]{>{\raggedleft\let\newline\\\arraybackslash\hspace{0pt}}m{#1}}

\begin{document}
%
%\frontmatter          % for the preliminaries
%
%\pagestyle{headings}  % switches on printing of running heads
%\addtocmark{Hamiltonian Mechanics} % additional mark in the TOC

%\tableofcontents
%
\mainmatter              % start of the contributions
\title{Discrete Gaussian Process Representations for Optimising UAV-based Precision Weed Mapping}
\titlerunning{Discrete Representations for UAV-based Weed Mapping}  % abbreviated title (for running head)
\author{Jacob Swindell\inst{1} \and Madeleine Darbyshire\inst{1},
Marija Popovi\'{c}\inst{2} \and Riccardo Polvara\inst{1}}
\authorrunning{Jacob Swindell et al.} % abbreviated author list (for running head)
\tocauthor{Jacob Swindell, Madeleine Darbyshire, Marija Popovi\'{c}, Riccardo Polvara}
\institute{Lincoln Centre for Autonomous Systems (L-CAS),\\ School of Engineering \& Physical Sciences, University of Lincoln, Lincoln, UK.
\and
MAVLab, Faculty of Aerospace Engineering, TU Delft, Delft, Netherlands}

\maketitle              % typeset the title of the contribution

\begin{abstract}
%
  %% WHY 
  % Use 1-2 not too long sentences, which clearly answer the WHY question: 
  % Why is this relevant, why should I care? Motivate why the stuff that 
  % you enable is relevant (not neccessarily equal to the technique)

  %% WHICH PROBLEM 
  % One sentence that explain the problem the paper addresses/ivestigates
  % Start with: In this paper, we address the problems of \dots

  %% HOW & WHAT
  % Around 3 sentences that explain how to approach the problem in general and answers:
  % How to solve the problem in general? (1/2 - 1 sentence)
  % What makes our approach special? What are we actually doing? What is new?
  
  %% IMPLEMENTATION, EVALUATION, WHAT FOLLOWS
  % 1-2 sentences what the experiments show and potentiall what follows from
  % your great work for the research community or the rest of the world ;-)

Accurate agricultural weed mapping using UAVs is crucial for precision farming applications. Traditional methods rely on orthomosaic stitching from rigid flight paths, which is computationally intensive and time-consuming. Gaussian Process (GP)-based mapping offers continuous modelling of the underlying variable (i.e. weed distribution) but requires discretisation for practical tasks like path planning or visualisation. Current implementations often default to quadtrees or gridmaps without systematically evaluating alternatives. This study compares five discretisation methods: quadtrees, wedgelets, top-down binary space partition (BSP) trees using least square error (LSE), bottom-up BSP trees using graph merging, and variable-resolution hexagonal grids. Evaluations on real-world weed distributions measure visual similarity, mean squared error (MSE), and computational efficiency. Results show quadtrees perform best overall, but alternatives excel in specific scenarios: hexagons or BSP LSE suit fields with large, dominant weed patches, while quadtrees are optimal for dispersed small-scale distributions. These findings highlight the need to tailor discretisation approaches to weed distribution patterns (patch size, density, coverage) rather than relying on default methods. By choosing representations based on the underlying distribution, we can improve mapping accuracy and efficiency for precision agriculture applications.

\end{abstract}
%
%%%%%%%%%%%%%%%%%%%%%%%%%%%%%%%%%%%%%%%%%%%%%%%%%%%%%%%%%%%%%%%%%%%%%%%%%%%%%%%%
\section{Introduction}
\label{sec:intro}

%%%%%%%%%%%%%%%%%%%
%% WHY: 
% First, answer the WHY question: Why is that relevant? Why should I be
% motivated to read the paper? Why should I care? (1 paragraph, 2-5 sentences)

% Unmanned aerial vehicles (UAVs) are commonly used for this purpose and can produce distribution maps using techniques such as Kriging with 2D Gaussian processes (GP).  

%For full-resolution orthomosaics using precise real-time kinematic (RTK) localisation flight time and processing time combined can take up to $\sim$10 hours \cite{pickett_small_2023}

\begin{figure}[h] % 'h' means place the figure here
    \vspace*{-0.5cm}
    \centering
    \includegraphics[width=\textwidth]{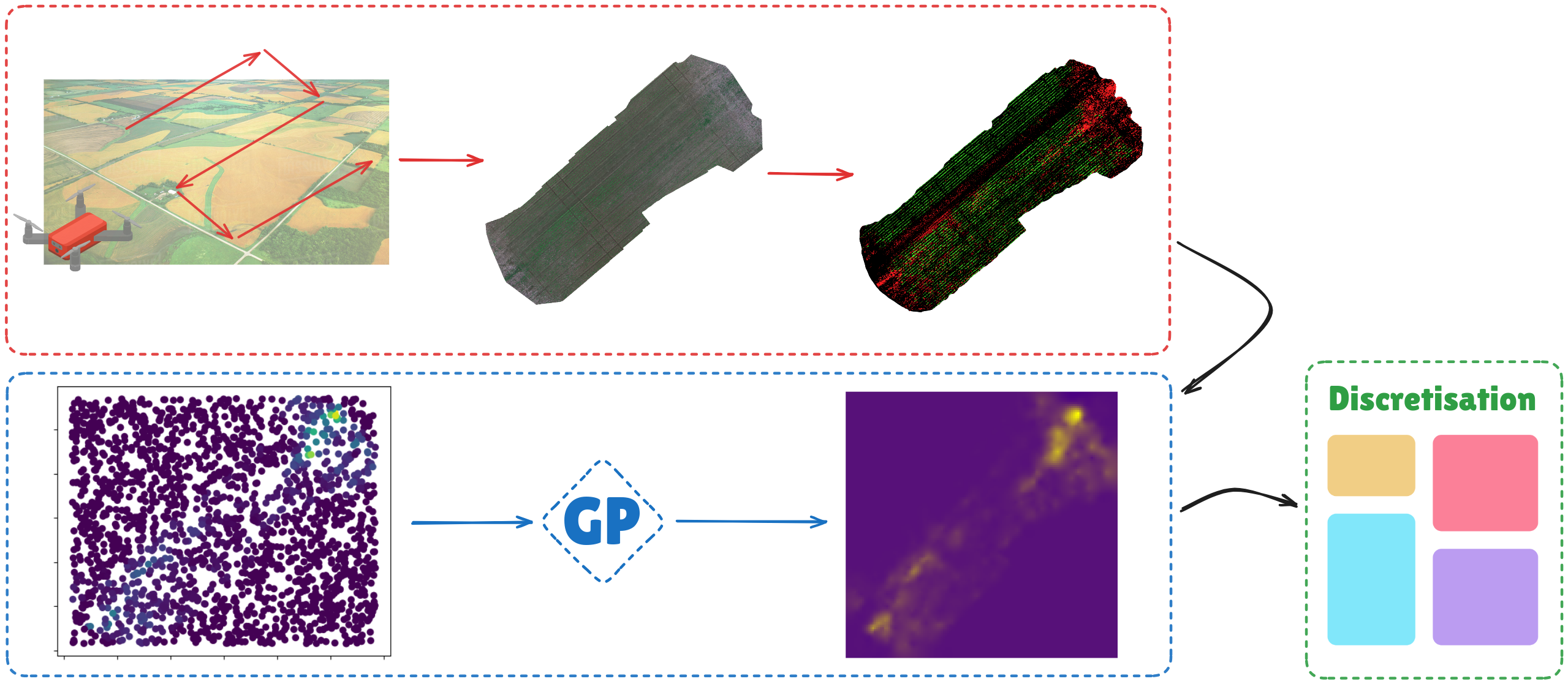} % Adjust width as needed
    \caption{Conventional weed mapping uses UAV orthomosaics and deep learning to label coverage, yet pixel-level segmentation may be superfluous. We show that a Gaussian process regressor, trained from uniform samples of the orthomosaic, can capture the underlying weed distribution, and compare multiple discretised representations for computational and statistical advantages.}
    \vspace*{-0.5cm}
    \label{fig:pipeline} % Reference this figure using \ref{fig:sample}
\end{figure}

Our project focuses on the production of accurate weed maps for precision agriculture tasks. Traditionally, unmanned aerial vehicles (UAVs) produce orthomosaic weed maps by stitching together overlapping images captured along a fixed lawnmower-like path, as seen at the top of Figure \ref{fig:pipeline}. The image processing procedure is typically performed offline and can be computationally intensive and time-consuming. Alternative survey strategies consider Gaussian process (GP) mapping~\cite{jin_adaptive-resolution_2022} which we focus on for the production of weed maps. While orthomosaics merge multiple overlapping images that ``cover" a field, GPs rely on recording spatially located measurement points and use probabilistic interpolation to estimate values in regions where data points are missing. There is no fixed requirement for the spatial arrangement of these data points, allowing GP mapping to accommodate arbitrary paths for data collection, which enables tasks such as adaptive path planning for efficient information gathering. In geostatistics, this technique is referred to as kriging, and it uses equations called variograms to characterise the spread of data for GP interpolations. It is commonly used to predict, map, and capture correlations of spatially distributed quantities, e.g. pollution, temperature, or weed pressure, while being computationally efficient enough to operate in real time on autonomous robots for small-scale datasets. Since 2D GPs store data continuously, they must first be discretised into a suitable representation. The choice of representation to discretise continuous GP-based weed maps significantly influences the fidelity with which the true GP state is maintained, as it is used to sample the mean and covariance at a given location of the GP rather than sampling the GP itself. This is because directly querying the GP requires $O(n^3)$ time for $n$ measurements. Representations that capture the true GP state more accurately enable greater reliability in tasks such as path planning without the time complexity of querying the GP.

%%%%%%%%%%%%%%%%%%%
%% WHICH PROBLEM
% Second, explain WHICH problem you are solving/address to solve.

%%%%%%%%%%%%%%%%%%%
%% HOW & WHAT
% Third, explain briefly how one can address the problem in general and mention 
% briefly what others/we before have done. Prepare the reader for your contribution 
% that comes in the next section (and not here!).

This paper evaluates different discrete representations for 2D GPs to improve the preservation of the GP state for spatial queries. Specifically, we focus on aerial weed mapping in UAV-based precision agriculture tasks. To address this problem, previous studies have used representations like quadtrees~\cite{samet_quadtree_1984} to discretise continuous GP weed maps. Current research~\cite{jin_adaptive-resolution_2022} focuses on improving the ability of these representations to capture the underlying data, without evaluating how alternative methods can be more suited to the target data distribution. This work explores the suitability of alternative structures for real-world weed mapping scenarios.

% Link to figure somewhere
% See \figref{fig:motivation} for an example.

% \begin{figure}[t]
%      \centering
%      \begin{subfigure}[b]{0.40\textwidth}
%          \centering
%          \includegraphics[width=\textwidth]{pics/000_reps/zstar.png}
%          \caption{Gridmap}
%          \label{fig:4_adjacency}
%      \end{subfigure}
%      % \hspace{0.5\textwidth}
%      \begin{subfigure}[b]{0.40\textwidth}
%          \centering
%         \includegraphics[width=\textwidth]{pics/000_reps/quadtree.png}
%         \caption{Quadtree}
%         \label{fig:8_adjacency}
%      \end{subfigure}
%     \caption{Gridmap and Quadtree representations for a Gaussian Process Weed Map}
%     \label{fig:4_vs_8_adjacency}
% \end{figure}

% \begin{figure}[t]
%      \centering
%      \begin{subfigure}[b]{0.90\textwidth}
%          \centering
%          \includegraphics[width=\textwidth]{pics/ortho.png}
%          % \caption{Baseline}
%          \label{fig:orthophoto}
%      \end{subfigure}
%      % \hspace{0.5\textwidth}
%      \begin{subfigure}[b]{0.90\textwidth}
%          \centering
%         \includegraphics[width=\textwidth]{pics/gp.png}
%         % \caption{Quadtree}
%         \label{fig:gp_regressor}
%      \end{subfigure}
%     \caption{A visual comparison between (top) the standard approach for weed mapping based on orthorectification of UAV images, and (both) our proposed methodology based on informative path planning and Gaussian process regressors.}
%     \label{fig:methodologies}
% \end{figure}

%%%%%%%%%%%%%%%%%%%
%% MAIN CONTRIBUTION & WHAT FOLLOWS FROM THAT
% Explain your contribution in one paragraph. This is a very important paragraph. 
% Always start that paragraph with: ``The main contribution of this paper is''
% and be SUPER-EXPLICIT for what you claim contribution (and maybe novelty) for.

The main contribution of this paper is a comprehensive evaluation of five discrete representations for 2D GP weed maps: quadtree, wedgelet~\cite{gao_application_2008}, binary space partition (BSP) with Least Square Error~\cite{radha_binary_1991}, BSP with region-based approach~\cite{salembier_binary_2000}, and a hexagonal grid~\cite{birch_rectangular_2007}. We compare these representations using various quality metrics, including visual similarity, mean squared error, and computational efficiency, by testing them on real-life weed distribution patterns captured by the WeedMap dataset~\cite{sa_weedmap_2018}.

%%%%%%%%%%%%%%%%%%%
%% OUR KEY CLAIMS (can be merged with the main contribution above if desired)
% Explicitly(!) state your claims in one (short) paragraph and make
% sure you pick them up again in the experiments and support every claim.

Our key claims are: (1) While quadtrees perform best on average across all metrics and maps, the performance of any given representation can vary widely depending on the underlying data distribution, specifically the number, size, variation, and coverage of weed patches, indicating that the map representation should be selected based on these distributional factors; (2) the BSP LSE approach is a promising alternative general representation as it achieves the best average mean squared error despite high computational costs.
These claims are backed up by the paper and our experimental evaluation.

%%%%%%%%%%%%%%%%%%%%%%%%%%%%%%%%%%%%%%%%%%%%%%%%%%%%%%%%%%%%%%%%%%%%%%%%%%%%%%%%
\section{Related Work}
\label{sec:related}

\textit{Orthomosaics} from UAV imagery can be generated through three main approaches: 2D mosaicking, structure from motion, and simultaneous localisation and mapping~\cite{zhang_aerial_2023}. In general, each method stitches together multiple smaller images to produce one large high-resolution image. This is typically performed offline due to high associated computational costs. The WeedMap~\cite{sa_weedmap_2018} dataset employs the 2D mosaic method, which combines drone localisation with image registration and feature matching between overlapping images. The overlap requirements vary by application, with examples ranging from 80\% front-lap/30\% side-lap~\cite{pena2015quantifying} to 60\% front-lap/30\% side-lap~\cite{lopez-granados_object-based_2016}, while the WeedMap dataset uses 80\% overlap in both directions. The process culminates in image fusion to create the orthomosaic, which is then georeferenced to assign pixelwise latitude and longitude coordinates.

% \paragraph{Real-Time Processing} was traditionally impossible to perform on the UAV due to limitations in computing power, thus machine learning models were performed offline. For example, a paper by \cite{castaldi_assessing_2017} utilized UAV-collected images to train a Support Vector Machine offline to produce a map informing herbicide patch spraying in maize. More modern UAV systems have started using real-time processing with UAVs using onboard GPUs to execute models on the fly. A paper by \cite{deng_lightweight_2020} used a combination approach, with a ground station laptop having computed the path for the drone and the AlexNet CNN semantic segmentation model performed by an onboard GPU. This system segments weeds from rice in a rice field mapping task. The downside of this approach is that smaller and faster models must be used as the UAV is expected to perform inference in real time; therefore, the segmentation quality will suffer compared to offline computation. There are many other papers in this area, such as a paper by \cite{de_camargo_optimized_2021} that also used onboard GPUs for semantic segmentation in identifying weeds in winter wheat. Additionally, a paper by \cite{menshchikov_real-time_2021} used real-time UAV detection for the poisonous hogweed plant.

\textit{Real-time processing} on UAVs has evolved from traditional offline processing due to computational constraints. In the past, algorithms were run offline as processing power was limited and could not be carried with small UAV payloads. Recently, embedded hardware has become more powerful and can incorporate onboard graphics processing units (GPUs) to enable real-time processing through parallelisation capabilities. Initially, offline methods, e.g. support vector machines (SVMs) were used for tasks such as herbicide patch spraying in maize \cite{castaldi_assessing_2017}. However, modern systems can leverage real-time processing by, for example, using a ground station laptop for path planning while the UAV uses an onboard semantic segmentation in rice fields \cite{deng_lightweight_2020}. While real-time processing requires smaller, faster models that may compromise segmentation quality, it has been successfully implemented in various applications, including weed identification in winter wheat \cite{de_camargo_optimized_2021} and poisonous hogweed plant detection \cite{menshchikov_real-time_2021}.

% \paragraph{Gaussian Processes} are starting to be used more in UAV mapping research in an attempt to move away from orthomosaics and their heavily constrained lawnmower-like paths. For example, a paper by \cite{jin_adaptive-resolution_2022} showcases a framework for UAV mapping with Gaussian processes to produce quadtree representations. This work details a method utilizing Gaussian process fusion to produce a quadtree map representation with adaptive resolution so that areas of high interest can be mapped with high resolution, while areas of low interest can be mapped with low resolution. The reasoning behind this representation provided in the paper is to improve the computational efficiency and compactness of the representation so that the UAV can interpret the map efficiently on the fly; an example of how this would be used is with dynamic path planning, which is demonstrated in the paper. This paper evaluates many other methods for producing quadtrees, evaluating their novel approach against them; it does not, however, compare against other representations. 

\textit{Gaussian Processes (GPs)} are emerging as an alternative to traditional orthomosaic mapping. Their advantages include being efficient enough to enable real-time terrain modelling, having built-in uncertainty measures, and supporting the use of non-standard surveying paths due to their ability to interpolate output data between registered data samples. For example, GPs have been used for UAV-based magnetic field mapping \cite{kuevor_fast_2023} using custom kernels and probabilistic sensor models to map the influence of magnetic fields. More recently, frameworks have been introduced to create adaptive resolution quadtree map representations using incrementally updated GP models via Bayesian filtering~\cite{jin_adaptive-resolution_2022}. In this, an integral kernel is used to calculate correlation over areas of the map rather than single points, which enables adaptive-resolution mapping in regions of interest. This approach aims to improve computational efficiency and map compactness for real-time UAV interpretation, which is beneficial for reactive path planning. While the study evaluates various quadtree production methods, it does not consider different mapping representations. Our paper addresses this gap by investigating alternative mapping structures, including binary space partitioning (BSP) trees and hexagon maps. GPs have also been used for UAV-based magnetic field mapping \cite{kuevor_fast_2023} using custom kernels and probabilistic sensor models to map the influence of magnetic fields. %We leverage these strengths of GPs for weed mapping using different map representations.

\textit{Image Abstractions} in computer vision vary in their ability to effectively maintain the underlying true data, as demonstrated by \cite{kassim_hierarchical_2009} who evaluated quadtrees, binary space partitioning (BSP) Trees, wedgelets, and quad-binary (QB) Trees, showing that BSP trees offer superior adaptability for representing shapes like diagonal lines compared to more rigid quadtree structures. BSP Tree applications include object tracking and occlusion \cite{caccavale_visual_2003}, collision detection \cite{ganter_dynamic_2021}, and image compression \cite{chopra_improved_2011}, with implementations following either top-down division or bottom-up merging approaches. Bottom-up BSP Trees have been used for segmentation by \cite{salembier_binary_2000} through recursive region merging and in UAV hyperspectral imaging \cite{veganzones_hyperspectral_2014} using watershed segmentation. Top-down approaches have utilised methods like the Hough transform for straight-line region detection \cite{radha_binary_1991} and line parameter computation for region division \cite{radha_image_1996}. Wedgelets, as outlined by \cite{gao_application_2008}, represent a modern evolution of wavelets, particularly excelling in preserving anisotropy in image features. Their effectiveness in compression, especially for animation and cartoons, is demonstrated by \cite{lee_2003_image}, while \cite{willett_fast_2004} show their utility in denoising medical images affected by Poisson noise. In terms of spatial representations, Hexagram Maps were evaluated by \cite{birch_rectangular_2007} for ecological applications, with benefits extending to UAV mapping and path planning. The hierarchical hexagonal grid system is applied in diverse fields, from tracking locust movements \cite{klein_application_2023} to urban mapping~\cite{wozniak_hex2vec_2021}, where it is used for creating semantic embeddings of city regions.

\textit{Our contribution} is a comprehensive evaluation of five discrete representations for GP weed maps: quadtrees, wedgelets, BSP Trees using least square error, BSP Trees using a region-based approach, and hexagonal grids. Our evaluation metrics include visual similarity, mean squared error, and computational efficiency computed based on various real-world weed distribution patterns. Our findings identify the most promising representations for general map distributions. We also prove, through spatial analysis, that certain representations are better suited to different map distribution patterns depending on the number, size, variation, and coverage of weed patches present in a field.

%%%%%%%%%%%%%%%%%%%%%%%%%%%%%%%%%%%%%%%%%%%%%%%%%%%%%%%%%%%%%%%%%%%%%%%%%%%%%%%%
\section{Methodology}

% The project aims to compare different unique representations against the standard Quadtrees and gridmaps. The comparison focuses on how accurately these representations capture the ground truth (grid map) from the GP, using five orthomosaics from the WeedMap dataset.

Our paper evaluates the use of discretised GP representations compared to an orthomosaic. The comparison focuses on how accurately these representations approximate the data distribution of the orthomosaic after training the GP.

\subsection{Dataset / Data processing}

% We use the WeedMap~\cite{sa_weedmap_2018} dataset to train the GP, based on which we produce the various discretised representations. 

The WeedMap~\cite{sa_weedmap_2018} dataset provides 5 semantically segmented orthomosaics of different sugar beet fields in Rheinbach, Germany. The semantic images use the labels Weed (Green), Crop (Red) and Background (Black). The orthomosaics provided by this dataset are used to train our GP, based on which we produce the various discretised representations. Only the ground-truth images are used for training, and we only focus on the red pixel labels for weeds, ignoring crop and background labels. 
% This data set was produced using a 2D orthomosaic method. 

% \begin{figure}[h]
%      \centering
%      \begin{subfigure}[b]{0.38\textwidth}
%          \centering
%          \includegraphics[width=\textwidth]{pics/Orthomosaic.png}
%          \caption{Orthomosaic}
%          \label{fig:orthomosaic}
%      \end{subfigure}
%      \begin{subfigure}[b]{0.53\textwidth}
%          \centering
%          \includegraphics[width=\textwidth]{pics/sampling.png}
%          \caption{Sampling}
%          \label{fig:sampling}
%      \end{subfigure}
%     \caption{Random Sampling with Average Pooling performed on WeedMap Orthomosaic}
%     \label{fig:sampling_orthomosaic}
% \end{figure}

\subsection{Continuous Weed Distribution}

Training the GP on the orthomosaic enables us to better approximate the weed distribution in the field. Many different GP libraries exist in Python. We chose PyKrige\footnote{\url{http://pykrige.readthedocs.io/}} due to its support of 2D kriging with many different default variogram models provided to model the data spread.

The GP requires scalar values that are spatially located via $(x,y)$ coordinates for training. To obtain training data for the GP from the orthomosaic map, we use random uniform sampling with an average pooling approach. We scatter random points across the orthomosaic and sample images of $150px\times150px$ centred around those points. From these small cropped regions, the average weed value is calculated from the number of red pixels (weeds) divided by the total number of pixels. 

% Usually a trained with data-driven methods to match the underlying distribution as closely as possible. We consider this out of scope for our paper, instead opting for choosing the best performing model from a set of standard variograms.

Training the GP requires selecting a variogram to model the similarity between data points. Usually, a variogram is trained with data-driven methods to match the underlying distribution as closely as possible~\cite{viana_disentangling_2013}. We consider this out of the scope of our paper, instead opting for choosing the best-performing model from a set of standard variograms. Table \ref{tbl:variogram_metrics} compares the Q1, Q2, and cR metrics of how different standard variogram models fit the sampled data.  

% \begin{table}[h]
%     \vspace*{-0.8cm}
%     \setlength{\tabcolsep}{8pt}
%     \caption{Variogram statistics}
%     \centering
%     \begin{tabular}{l|c|c|c}
%         Variogram Model & Q1 ($\approx$ 0) & Q2 ($\approx$ 1) & cR ($\approx$ 0) \\
%         \hline\hline
%         Hole-Effect & \textbf{0.0102} & 1.206 & \textbf{0.0005} \\ 
%         Exponential & 0.0124 & 1.776 & 0.0005 \\ 
%         Spherical & 0.0108 & \textbf{1.008} & 0.0011 \\ 
%         Linear & 0.0154 & 0.872 & 0.0014 \\ 
%         Power & 0.0216 & 0.915 & 0.0018 \\ 
%         Gaussian & 0.0266 & 1.364 & 0.0024 \\ 
%         % \multicolumn{4}{c}{\footnotesize{Note: Q1 and cR should be near 0, and Q2 near 1.}}
%     \end{tabular}
%     \vspace*{-0.5cm}
%     \label{tbl:variogram_metrics}
% \end{table}

\begin{table}[h]
    \vspace*{-0.8cm}
    \setlength{\tabcolsep}{8pt}
    \caption{Variogram statistics. Q1 measures the average error of the model predictions compared to the real data; a value close to 0 indicates very accurate predictions. Q2 measures if the model errors are consistent with the errors expected; a value close to 1 indicates a reliably accurate model. cR measures the average magnitude of the errors; a low cR shows the errors are small relative to the data scale.}
    \centering
    \setlength{\tabcolsep}{5pt}
    \begin{tabular}{l|c|c|c|c|c|c}
         & Hole-Effect & Exponential & Spherical & Linear & Power & Gaussian \\
        \hline\hline
        Q1 ($\approx$ 0) & \textbf{0.0102} & 0.0124 & 0.0108         & 0.0154 & 0.0216 & 0.0266 \\ 
        Q2 ($\approx$ 1) & 1.206           & 1.776  & \textbf{1.008} & 0.872  & 0.915  & 1.364 \\ 
        cR ($\approx$ 0) & \textbf{0.0005} & 0.0005 & 0.0011         & 0.0014 & 0.0018 & 0.0024 \\ 
        % \multicolumn{4}{c}{\footnotesize{Note: Q1 and cR should be near 0, and Q2 near 1.}}
    \end{tabular}
    \vspace*{-0.5cm}
    \label{tbl:variogram_metrics}
\end{table}

Although Table \ref{tbl:variogram_metrics} shows that the hole-effect variogram produces the best results, due to the non-standard properties of this variogram, we chose to use the exponential variogram instead. Most variogram model's covariance decreases with greater \emph{lag} distances (distance from datapoints); however, the hole-effect variogram is unique in that it exhibits an oscillatory behaviour where covariance periodically decreases, then increases with greater lag distances~\cite{pyrcz2003whole}. These oscillations do not map well to naturally occurring distributions such as weeds and can lead to unstable or ill-defined matrices during the kriging process, which increases the risk of numerical errors during training.

% \begin{figure}[h]
%      \centering
%      \begin{subfigure}[b]{0.3\textwidth}
%          \centering
%          \includegraphics[width=\textwidth]{pics/hole_effect.png}
%          \caption{Hole-Effect}
%          \label{fig:hole_effect}
%      \end{subfigure}
%      \begin{subfigure}[b]{0.3\textwidth}
%          \centering
%          \includegraphics[width=\textwidth]{pics/exponential.png}
%          \caption{Exponential}
%          \label{fig:exponential}
%      \end{subfigure}
%      \begin{subfigure}[b]{0.3\textwidth}
%          \centering
%          \includegraphics[width=\textwidth]{pics/spherical.png}
%          \caption{Spherical}
%          \label{fig:spherical}
%      \end{subfigure}
%      \begin{subfigure}[b]{0.3\textwidth}
%          \centering
%          \includegraphics[width=\textwidth]{pics/linear.png}
%          \caption{Linear}
%          \label{fig:linear}
%      \end{subfigure}
%      \begin{subfigure}[b]{0.3\textwidth}
%          \centering
%          \includegraphics[width=\textwidth]{pics/power.png}
%          \caption{Power}
%          \label{fig:power}
%      \end{subfigure}
%      \begin{subfigure}[b]{0.3\textwidth}
%          \centering
%          \includegraphics[width=\textwidth]{pics/Gaussian.png}
%          \caption{Gaussian}
%          \label{fig:Gaussian}
%      \end{subfigure}
%         \caption{Gridmaps produced from each Variogram Model}
%         \label{fig:variogram_gridmaps}
% \end{figure}

\subsection{Discrete GP Representations}

Next, we briefly describe the steps required to construct each representation. We chose all methods based on their common use in the literature for computer vision~\cite{kassim_hierarchical_2009}. Our contribution is to evaluate these representations on 2D GP weed maps to discover the benefits they provide over gridmaps and quadtrees used in current literature.

\textit{BSP Least Squared Error (LSE) Trees}~\cite{radha_binary_1991} work by recursively dividing an image into two homogeneous regions. We first define a region using corner points of an $n$-gon and then establish parameter domains based on line equations. We sample possible dividing lines, pruning them using the LSE Partitioning Line (LPL) transform thresholds~\cite{radha_fast_1991}, and select the best line based on mean squared error calculations. This recursive process continues until either the max depth (9) or an arbitrarily chosen homogeneity criterion ($2\times10^{-4}$) is reached.

\textit{Hexagon maps}~\cite{birch_rectangular_2007} produce a variable-resolution based on minimising error. Starting with a uniform grid map, we convert coordinate cells to hexagons. We calculate multi-resolution parents by progressively increasing resolution and computing the average ratio of weed to background values and mean squared error for each hexagon. The final representation is created by selecting hexagons based on error thresholds, reflecting the region size and homogeneity.

\textit{BSP Tree Region}~\cite{salembier_binary_2000} approach differs from the LSE method by using a bottom-up strategy instead of a top-down one. Beginning with a grid map converted to a 4-adjacency graph, the Kruskal algorithm~\cite{najman_playing_2013} constructs an altitude-ordered binary partition tree by merging nodes based on edge weights. The tree is then pruned by removing and merging subtrees with children nodes that occupy $<10$ pixels, setting the minimum region size for the BSP tree. We calculate the average weed values for the filtered regions and display them to produce the final image representation.

\textit{Wedgelets}~\cite{gao_application_2008} combine aspects of quadtree and line-division approaches. It starts with standard quadtree recursion but adds the step of checking for optimal dividing lines within regions. When a region is not homogeneous, we evaluate various line orientations and apply a threshold to determine whether a line can adequately represent the region. Unlike the BSP LSE approach, wedgelet works with square areas and uses threshold-based line selection rather than LSE calculations.

\subsection{Evaluation Metrics}
\label{subsec:eval_metrics}

To evaluate the information loss for each representation, we compute the following metrics by comparing them against the original GP. We use a high-resolution gridmap to represent the continuous distribution as closely as possible.
\begin{enumerate}
\item \textbf{Structural Similarity Index Measure (SSIM)}: Evaluates image similarity by comparing luminance, contrast, and structure, mimicking human visual perception.
\item \textbf{Hamming Distance (HD)}: A perceptual hash captures the visual features and represents them as a string of letters and numbers. Visually similar images have similar hashes, resulting in smaller Hamming distances.
\item \textbf{Mean Squared Error (MSE)}: Measures pixelwise differences between images, focusing on computational accuracy rather than visual similarity.
\end{enumerate}
We also measure execution time and memory usage for each representation to assess their efficiency and suitability for real-time UAV applications. Our experiments were conducted on an AMD Ryzen 7 7700 CPU (3.8 GHz, 8 cores) and 32 GB DDR5 RAM. 

% \begin{enumerate}
% \item \textbf{Time (Seconds)}: How long does it take to construct a given representation.
% \item \textbf{Space (Mb)}: How much space does the representation take up in memory. 
% \end{enumerate}

% \subsection{Testing Protocol}

% To account for stochastic elements in the implementations and sampling processes, metrics are calculated multiple times per orthomosaic. The mean and standard deviation of these measurements provide robust performance indicators and measure representation quality variability.

%%% Move this to underneath tables /\ /\ /\ /\

%% Describe your approach. It is okay to divide the main section
%%  into a few subsections (e.g., 2-4 subsections).

%%%%%%%%%%%%%%%%%%%%%%%%%%%%%%%%%%%%%%%%%%%%%%%%%%%%%%%%%%%%%%%%%%%%%%%%%%%%%%%%
\section{Experimental Evaluation}
\label{sec:exp}

%% Repeat the main focus/objective with one single(!) sentence starting with:
%
Our key aim is to compare the effectiveness of five different discretised GP representations for UAV field mapping.
%
%% Explain the reader that the experiments with support all claims
%% (same list as in the intro!) starting the paragraph with:}
%
We present our experiments to show how suitable each representation is to a specific map given the key metrics in Sec. \ref{subsec:eval_metrics}. Our findings expose the benefits of different map representations for specific distributions. 

The results of our experiments support our key claims, which are: (1) while quadtrees perform best on average across all metrics and maps, the performance of any given representation can vary widely depending on the underlying data distribution, specifically the number, size, variation, and coverage of weed patches, indicating that the map representation should be selected based on these distributional factors; (2) the BSP LSE approach is a promising general-use representation to use without knowledge of map distribution as it achieves the best average mean squared error despite high computational costs.

% \subsection{Experimental Setup}

%% If needed (and only then!) say also a few words about the experimental
%% setup, the datasets, and used parameters. You can use a separate subsection if you
%% want to put the focus on that but often that is not needed.}

%% Note 1: It MUST be always crystal clear (a) WHY an experiment is there
%% (e.g., to support a claim, to show that the approaches useful for real-word
%% systems, to show the performance, or to provide a baseline comparison), (b)
%% WHAT it wants to show (which claim/property exactly), and (c) HOW it aims at 
%% showing this. This is ESSENTIAL for a good evaluation. Think about when BEFORE
%% designing an experiment.  IMPORTANT: Every experiment MUST start with something 
%% like:  The next experiment is presented to show \dots and thus for supporting our 
%% first claim.

%% Note 2: Start with the most important/impressive experiment first. Make
%% his a key story of the paper. Keep the order of the claims, i.e., re-order
%% claims in the intro/before if needed. 

%%%%%%%%%%%%%%%%%%%%%%%%

\subsection{Qualitative Results}

\begin{figure}[h]
     \vspace*{-0.5cm}
     \centering
     \begin{subfigure}[b]{0.16\textwidth}
         \centering
         \includegraphics[width=\textwidth]{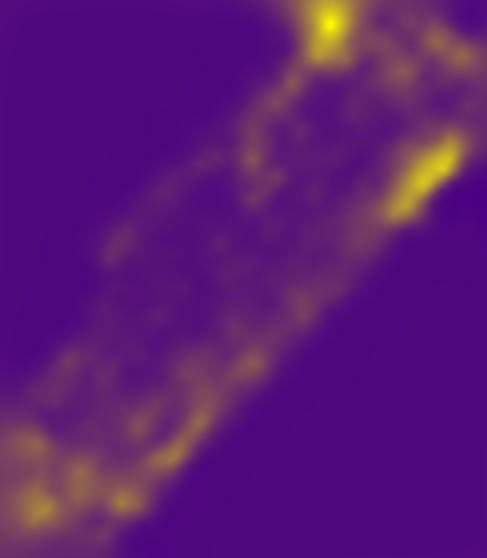}
         \caption{GP-Gridmap}
         \label{fig:000_gridmap}
     \end{subfigure}
     \begin{subfigure}[b]{0.16\textwidth}
         \centering
         \includegraphics[width=\textwidth]{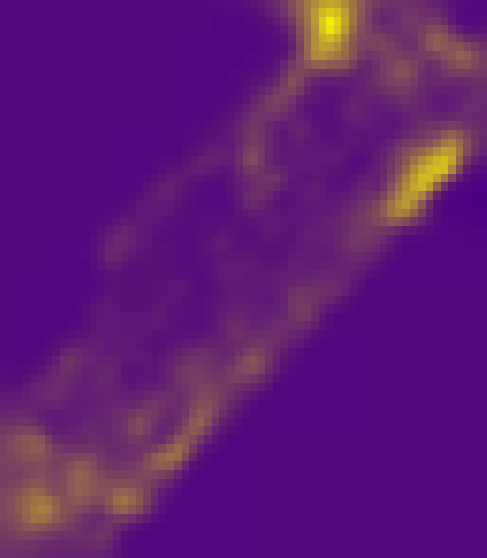}
         \caption{Quadtree}
         \label{fig:000_quadtree}
     \end{subfigure}
     \begin{subfigure}[b]{0.16\textwidth}
         \centering
         \includegraphics[width=\textwidth]{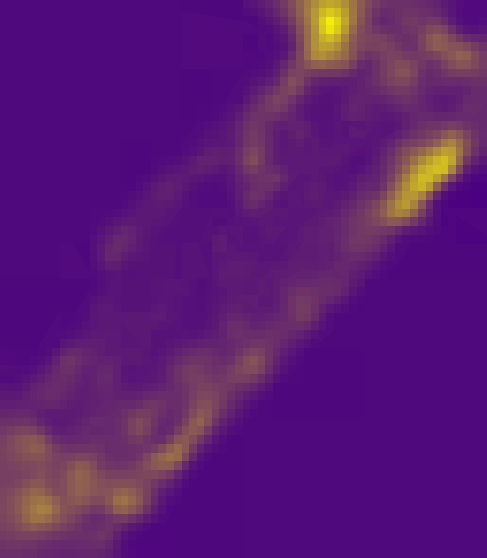}
         \caption{Wedgelet}
         \label{fig:000_wedgelet}
     \end{subfigure}
     \begin{subfigure}[b]{0.16\textwidth}
         \centering
         \includegraphics[width=\textwidth]{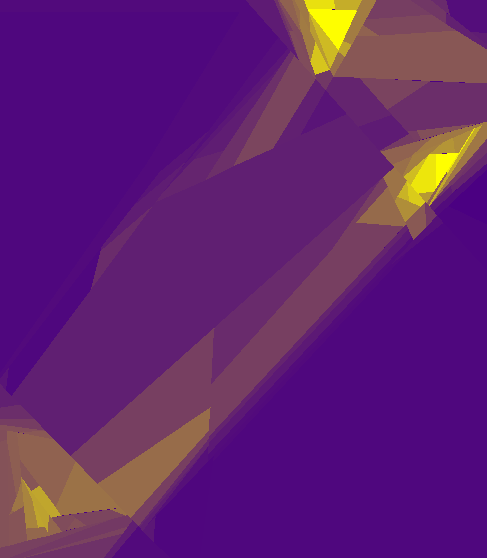}
         \caption{BSP LSE}
         \label{fig:000_bsp}
     \end{subfigure}
     \begin{subfigure}[b]{0.16\textwidth}
         \centering
         \includegraphics[width=\textwidth]{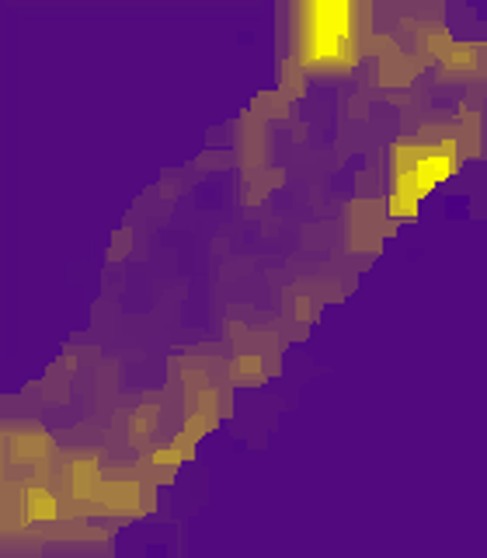}
         \caption{BSP Region}
         \label{fig:000_bsp_r}
     \end{subfigure}
     \begin{subfigure}[b]{0.16\textwidth}
         \centering
         \includegraphics[width=\textwidth]{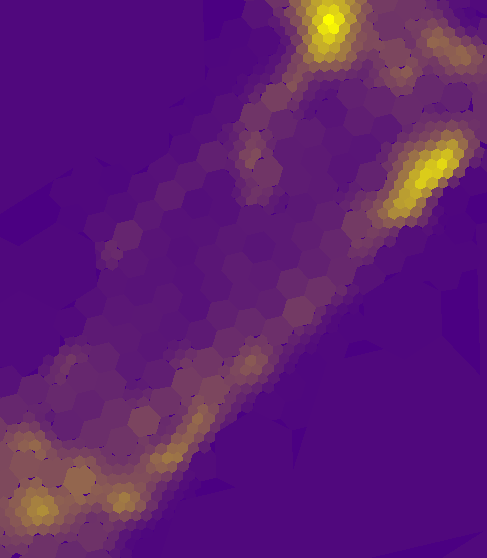}
         \caption{Hexagon}
         \label{fig:000_h3}
     \end{subfigure}
    \caption{Representations for 000\_gt orthomosaic in the WeedMap dataset. Bright spots show a high weed presence while dark spots show low weed presence. }
    \vspace*{-0.5cm}
    \label{fig:all_orthomosaics}
\end{figure}

We select the first orthomosaic from the WeedMap dataset (000\_gt.png) to train the GP and create a high-resolution gridmap, which is then used to compute the different representations. Figure \ref{fig:all_orthomosaics} shows the degree to which they visually capture the high-resolution grid map shown in Figure \ref{fig:000_gridmap}. While BSP LSE (Figure \ref{fig:000_bsp}) appears most abstract, it efficiently represents hotspots with concentrated partitions and fewer partitions in sparse areas. Quadtree (Figure \ref{fig:000_quadtree}) and wedgelet (Figure \ref{fig:000_wedgelet}) are visually similar and the closest to the original grid map, followed by the hexagon representation (Figure \ref{fig:000_h3}). Additionally, the BSP region representation (Figure \ref{fig:000_bsp_r}) has sharp, distinct regional boundaries that accurately show the high weed concentration in that portion of the map.

\subsection{Quantitative Results}

Table \ref{tbl:representation_simularity_metrics} shows the mean and standard deviation of each metric defined in Section \ref{subsec:eval_metrics} for each representation.

\begin{table}[ht]
    \vspace*{-0.5cm}
    \caption{Similarity metrics results for each orthomosaic (0-4). Structural similarity compares luminance, contrast, and structure. Hamming distance compares a perceptual hash of the images. Mean squared error measures pixelwise differences between images. Results are shown in the format "mean(std dev)" across 10 trials. Best results in bold. Lower is better for all metrics. }
    \centering
    {\small
    \setlength{\tabcolsep}{1.7pt}
    \begin{tabular}{c|c|c|c|c|c|c}
        & \#Map & Quadtree & Wedgelet & BSP LSE & BSP Region & Hexagon \\
        \hline\hline
        \parbox[t]{4mm}{\multirow{5}{*}{\rotatebox[origin=c]{90}{1 - SS (e-04)}}} 
        & 000 & \textbf{0.23(0.13)} & 0.28(0.14) & 4.69(3.19) & 3.91(5.38) & 0.45(0.17) \\
        & 001 & \textbf{0.20(0.11)} & 0.19(0.12) & 4.05(1.66) & 1.49(0.28) & 0.51(0.10) \\
        & 002 & 3.20(6.58) & 3.04(6.04) & 10.01(15.01) & 3.71(2.06) & \textbf{2.38(2.34)} \\
        & 003 & \textbf{1.79(1.54)} & 1.84(1.55) & 4.20(1.30) & 3.27(1.32) & 12.10(3.65) \\
        & 004 & 12.44(12.76) & 13.19(12.05) & \textbf{7.27(9.07)} & 34.94(33.76) & 15.39(11.46) \\
        \hline
        \parbox[t]{4mm}{\multirow{5}{*}{\rotatebox[origin=c]{90}{HD}}} 
        & 000 & \textbf{566(130.16)} & 799.43(162.34) & 1762(99.48) & 1323.71(187.03) & 1163.43(200.55) \\
        & 001 & \textbf{660.25(134.05)} & 777(157.72) & 1806.25(95.09) & 1390.25(98.04) & 1428(129.20) \\
        & 002 & \textbf{1055.5(101.09)} & 1257.25(96.52) & 1938(64.33) & 1641(152.60) & 1755.75(111.78) \\
        & 003 & \textbf{441.5(46.34)} & 590.5(108.63) & 1705.75(60.86) & 1233.75(66.65) & 1883.75(64.99) \\
        & 004 & \textbf{560(127.61)} & 676.86(104.82) & 1681.43(37.70) & 1640(235.53) & 2071.43(110.73) \\
        \hline 
        \parbox[t]{4mm}{\multirow{5}{*}{\rotatebox[origin=c]{90}{MSE}}} 
        & 000 & \textbf{101.82(46.11)} & 103.28(45.46) & 102.75(27.11) & 129.08(31.04) & 115.89(32.69) \\
        & 001 & 110.71(33.24) & 110.96(32.52) & 91.39(26.53) & \textbf{89.85(7.46)} & 111.44(19.65) \\
        & 002 & \textbf{75.57(5.17)} & 78.03(5.04) & 114.88(26.86) & 107.93(29.85) & 108.93(31.65) \\
        & 003 & 114(27.62) & 114.47(26.63) & 108.29(38.10) & 123.73(30.25) & \textbf{97.85(17.47)} \\
        & 004 & 146.52(23.97) & 141.38(21.48) & 126.56(19.07) & 127.14(10.62) & \textbf{123.64(17.45)} \\
        % \multicolumn{7}{c}{\footnotesize{Note: Lower is better for all metrics}}
    \end{tabular}
    }
    \vspace*{-0.5cm}
    \label{tbl:representation_simularity_metrics}
\end{table}

Table \ref{tbl:representation_simularity_metrics_mean_std_of_mean} shows the mean and standard deviation across 10 trials of the metrics for all 5 orthomosaics. It indicates that the quadtree performs best on average across the 5 orthomosaic maps for SSIM and HD. However, considering the MSE, the BSP LSE approach performs slightly better than the quadtree and wedgelet representations. Since the orthomosaics tested were captured at a roughly 45 degree tilt, it is possible that the BSP LSE can better represent these diagonal regions due to its ability to divide regions at arbitrary angles. The BSP LSE approach has the lowest standard deviation, demonstrating a more consistently well-performing representation.

\begin{table}[ht]
    % \vspace*{-0.5cm}
    \captionsetup{justification=centering}
    \caption{Similarity metrics results averaging all orthomosaics. SS is represented as 1-SS (e-04). Lower is better for all metrics.}
    \centering
    {\fontsize{8.4pt}{10pt}\selectfont
    \setlength{\tabcolsep}{4pt}
    \begin{tabular}{l|c|c|c|c|c}
        & Quadtree & Wedgelet & BSP LSE & BSP Region & Hexagon \\
        \hline\hline
        SS& \textbf{3.57(5.11)} & 3.71(5.43) & 6.04(2.57) & 9.47(14.27) & 6.16(7.06) \\
        HD & \textbf{656.65(236.08)} & 820.21(258.18) & 1778.69(101.49) & 1445.74(186.26) & 1660.47(363.63) \\
        MSE & 109.73(25.51) & 109.62(22.76) & \textbf{108.78(13.16)} & 115.55(16.60) & 111.55(9.48) \\
        % \multicolumn{6}{c}{\footnotesize{Note: Lower is better for all metrics}}
    \end{tabular}
    }
    % \vspace*{-0.5cm}
    \label{tbl:representation_simularity_metrics_mean_std_of_mean}
\end{table}

\subsection{Computational and Memory Performance}

Table \ref{tbl:representation_time_space_mean_std_of_mean} shows the mean and standard deviation of the execution time and memory usage for each representation, run 10 times across the 5 orthomosaics using the hardware specified in Sec. \ref{subsec:eval_metrics}. Our findings broadly overview how each representation performs across each map. This table additionally shows the memory consumption of the grid map used to train the representations. The BSP region approach is the most computationally efficient, while the hexagon approach consumes the least memory. 

\begin{table}[ht]
    \vspace*{-0.5cm}
    \caption{Time and Space metrics. Lower is better for all metrics.}
    \centering
    {\small
    \setlength{\tabcolsep}{4pt}
    \begin{tabular}{l|c|c|c|c|c|c}
        & Quadtree & Wedgelet & BSP LSE & BSP Region & Hexagon & Grid map\\
        \hline\hline
        Time (s) & 0.02(0.01) & 4.13(1.29) & 141(68) & \textbf{0.01(0.00)} & 7.32(2.02) & N/A \\
        Space (Mb) & 0.77(0.20) & 0.72(0.28) & 0.23(0.08) & 1.02(0.27) & \textbf{0.05(0.03)} & 11.94(3.14) \\
        % \multicolumn{7}{c}{\footnotesize{Note: Lower is better for all metrics}}
    \end{tabular}
    }
    \vspace*{-0.5cm}
    \label{tbl:representation_time_space_mean_std_of_mean}
\end{table}

\subsection{Correlation}

% High positive global moran's I shows strong clustering of similar values together (High and Low values), negative moran's I shows disimmilar values commonly found together, High with Low, 0 shows random distribution and now clustering
% HH/LL ratio show more high clusters than low clusers or vice-verca
% HL/LH ratio show if anomolies are primarily high value in low-value regions or low value in high value region

The purpose of our final experiment is to investigate the relationship between selected field features (e.g. weed coverage ratio, number of weed patches) computed by analysing the RGB ortophotos of the five fields, and the performance of different partitioning methods. The performance metrics used are the same from the previous section - namely 1-SS, HD, and MSE - all of which benefit from being as low as possible. To quantify how strongly each feature relates to each metrics, we compute pairwise correlations using Spearman's rank correlation.

\begin{table}[ht]
\vspace*{-0.8cm}
\centering
\caption{Spearman's correlation values between features and performances metrics. We report the max and min values along with their associated performance metric.}
% \resizebox{\textwidth}{!}{
\begin{tabular}{l @{\hspace{0.5cm}} cc @{\hspace{0.5cm}} cc}
\hline
\textbf{Feature} & \multicolumn{4}{c}{\textbf{Correlation}}\\
\cline{2-3}\cline{4-5}
& \textbf{Metric} & \textbf{Max Value} & \textbf{Metric} & \textbf{Min Value} \\
\hline\hline
weed\_coverage\_ratio & Hex\_MSE           & 0.50  & Quadtree\_1-SS   & -0.70 \\
weed\_patches         & BSP\_LSE\_HD      & 0.70  & Quadtree\_1-SS   & -0.90 \\
largest\_patch\_size & Quadtree\_1-SS & 0.90  & BSP\_LSE\_HD     & -0.30 \\
avg\_patch\_size       & Hex\_MSE           & 0.60  & BSP\_LSE\_1-SS   & -0.60 \\
patch\_size\_std      & Hex\_MSE           & 1.00  & BSP\_LSE\_HD     & -0.40 \\
% grid\_mean\_coverage       & H3\_MSE           & 0.5000  & Quadtree\_1-SS   & -0.7000 \\
% grid\_std\_coverage        & H3\_MSE           & 0.6000  & Quadtree\_1-SS   & -0.6000 \\
% grid\_max\_coverage        & H3\_MSE           & 0.6000  & Quadtree\_1-SS   & -0.6000 \\
dbscan\_num\_clusters      & BSP\_LSE\_HD      & 0.70  & Quadtree\_1-SS   & -0.90 \\
dbscan\_avg\_cluster\_size & Hex\_MSE           & 0.60  & BSP\_LSE\_1-SS   & -0.60 \\
global\_autocorrelation & Hex\_MSE           & 0.70  & Hex\_1-SS         & -0.70 \\
hotspot\_to\_coldspot\_ratio & Quadtree\_HD    & 1.00  & Hex\_MSE           & -1.00 \\
hot\_to\_cold\_outlier\_ratio & BSP\_Region\_HD  & 0.90  & BSP\_Region\_MSE   & -0.10 \\
\hline
\end{tabular}
\vspace*{-0.5cm}
% }
\label{tab:pos_neg_corr}
\end{table}

Our findings are reported in Table~\ref{tab:pos_neg_corr}. They indicate that quadtree tends to yield improved (lower) 1-SS in settings where multiple, relatively small patches are distributed, but deteriorates if there is a single dominant patch. In contrast, Hex map's MSE rises with increasing patch size or variability, suggesting potential limitations in highly heterogeneous fields. BSP LSE's HD worsens in scenarios featuring numerous disjoint patches, while its 1-SS can be relatively low when the average patch size is large. Additionally, BSP region's MSE performs best when outliers are primarily low values in high-value regions, which would show if a distribution is characterised by large high value patches containing many small low-value patches within. Consequently, a positive correlation for a given feature suggests that higher values of that feature are associated with a worsening of the relevant performance metric (and vice versa for negative correlations). These results underscore the need to match each method to the specific weed distribution context to achieving optimal discretisation accuracy.

\subsection{Discussion}

% Finally, we analyze out methods with respect to the ability to \dots (backup the last claim)

%% Briefly summarize the evaluation and what follows with approx. 2 sentences.
% In summary, our evaluation suggests that our method
% provides competitive \dots At the same time, our method is fast enough
% for online processing and has small memory demands. Thus, we supported
% all our claims with this experimental evaluation.

The quadtree representation achieves the highest mean scores across all metrics and maps, as seen in Table \ref{tbl:representation_simularity_metrics}, particularly excelling in Hamming distance measurements. However, specific representations show better performance in different distribution conditions. BSP region-based and BSP LSE approaches performed best for map 001's MSE, which is characterised by high weed coverage with extreme cluster fragmentation, but large weed cluster size. Hexagon maps excelled in maps 003 and 004's MSE, both with moderate weed coverage and many smaller clusters. BSP LSE achieved the best SSIM score for map 004, which contains many smaller clusters and has the largest individual cluster of any map. Quadtrees exhibit higher standard deviations, while other representations demonstrate more consistent performance metrics.

Table \ref{tbl:representation_time_space_mean_std_of_mean} shows that the quadtree and BSP region approaches are the most computationally efficient, while BSP LSE is the least, taking more than 2 minutes per representation\cite{j_time_2012} which is too high for real-time UAV mapping. However, BSP LSE shows excellent memory efficiency, using less than half the storage of quadtree and wedgelet representations. The hexagon representation is highly memory-efficient due to its unique indexing system, while all representations achieved significant compression compared to the original grid map.

We find that quadtrees are the best general-purpose solution, and are particularly effective with numerous but moderately-sized weed patches. However, we also discover that alternative representations offer advantages in specific scenarios, e.g. BSP LSE and hexagon maps perform well with larger, more dominant patches. The BSP LSE approach, despite its computational intensity, obtains the best mean MSE and has superior memory efficiency. This suggests its potential for future research and optimisation as an alternative standard representation for GP-based mapping.

Table~\ref{tab:method_guidelines} summarises our recommended representation choice for new fields, based on our correlation analysis of patch properties and method performance in Table~\ref{tab:pos_neg_corr}. Notably, quadtree tends to perform better with numerous, smaller patches, while both BSP-LSE and hexagon maps excel in fields with large patch sizes.

\begin{table}[ht]
\vspace*{-0.5cm}
\centering
\caption{Guidelines for selecting a segmentation/partitioning method (Quadtree, Hex, or BSP LSE, BSP Region) based on weed distribution conditions.}
\resizebox{\textwidth}{!}{
\begin{tabular}{p{5.5cm} p{9cm}}
\hline
\textbf{Condition} & \textbf{Guideline} \\
\hline\hline
\textbf{Many smaller patches} 
& \textbf{Quadtree}: Particularly effective when weed patches are numerous but moderately sized. \\
\textbf{One large (dominant) patch} 
& \textbf{Hex or BSP LSE}: Quadtree performance may deteriorate if a single patch dominates. \\
\textbf{Highly variable patch size} 
& \textbf{BSP LSE or Quadtree}: Hex’s MSE can rise steeply with increasing patch size variability. \\
\textbf{Many disjoint weed patches/clusters} 
& \textbf{Quadtree}: Often handles a high count of smaller patches better; BSP LSE’s HD grows under many disjoint patches. \\
\textbf{Patches are large on average} 
& \textbf{BSP LSE}: Tends to produce lower 1--SS when patch sizes are consistently large. \\
\hline
\end{tabular}
}
\vspace*{-0.5cm}
\label{tab:method_guidelines}
\end{table}

%%%%%%%%%%%%%%%%%%%%%%%%%%%%%%%%%%%%%%%%%%%%%%%%%%%%%%%%%%%%%%%%%%%%%%%%%%%%%%%%
\section{Conclusion}
\label{sec:conclusion}

This manuscript evaluated alternative discredited representations to quadtrees for Gaussian process-based mapping, including BSP LSE, BSP region, wedgelet, and hexagon maps. Our experimental results demonstrate that these alternatives can provide more accurate representations for specific weed distributions depending on the number, size, variation, and coverage of weeds compared to traditional grid maps and quadtrees while maintaining strong compression rates.

We provide a guideline through spatial analysis of what representations should be investigated depending on the distribution of the underlying data. In general, for large dominant patches, hexagon maps or BSP LSE representations should be investigated for use as they proved the most efficient for this distribution pattern, however for distributions comprised of many dispersed regions, quadtrees provide the most benefit. 

Future work will evaluate these representations using more diverse datasets and investigate their applicability for online UAV-based mapping and planning.

% The representations evaluated in this paper were only tested on a single dataset comprising of weed maps. While we have made an attempt to generalise for different types of distributions, additional testing is required in more diverse datasets to confidently say in which scenario each representation should be considered.

% Further research is needed to evaluate these representations' effectiveness for online mapping applications, as the ability to make use of the different representation data structures in online planning algorithms remains untested.

% The project concludes that while Quadtrees serve well as a general-purpose solution, researchers should consider alternative representations that might better suit their specific mapping tasks rather than over-optimizing Quadtree implementations.

%%%%%%%%%%%%%%%%%%%%%%%%%%%%%%%%%%%%%%%%%%%%%%%%%%%%%%%%%%%%%%%%%%%%%%%%%%%%%%%%
%% Future work: Use only if applicable -- but if so, use the following
%% sentence to start:
% Despite these encouraging results, there is further space for improvements. 
%% In general, I avoid explaining future work in 6-8 page conference papers...

\section*{Acknowledgments}
This work was supported by Engineering and Physical Sciences Research Council, UK Projects GAIA (Ref: EP/Y003438/1) and AgriFoRwArdS (Ref: EP/S023917/1).% <-this % stops a space
%%%%%%%%%%%%%%%%%%%%%%%%%%%%%%%%%%%%%%%%%%%%%%%%%%%%%%%%%%%%%%%%%%%%%%%%%%%%%%%%
% Only if applicable
%\section*{Acknowledgments}
%We thank XXX for fruitful discussions and for \dots

\bibliographystyle{plain_abbrv}

% All new citations should go to new.bib. The file glorified.bib should go
% be the one from the ipb server. After paper or related work has been
% written merge the entries from new.bib to glorified.bib ON THE SERVER,
% replace the glorified.bib in this repository and empty the new.bib
\bibliography{glorified,new}

\end{document}